\documentclass[pageno]{jpaper}

\usepackage[normalem]{ulem}
\usepackage{graphicx}
\usepackage{grffile}
\usepackage{amsmath,amsfonts,amssymb}
\usepackage{multicol,multirow}
\usepackage{stfloats}
\usepackage{makecell}
\usepackage[numbers,sort]{natbib}

\begin{document}

\title{
Deep Learning Inference in Facebook Data Centers: Characterization, Performance Optimizations and Hardware Implications}

\date{}
\author{
Jongsoo Park\thanks{jongsoo@fb.com $^\dagger$mnaumov@fb.com},\ \ 
Maxim Naumov$^\dagger$,\ \ 
Protonu Basu,
Summer Deng,
Aravind Kalaiah,
Daya Khudia,
James Law, \and
Parth Malani,
Andrey Malevich,
Satish Nadathur,
Juan Pino,
Martin Schatz,
Alexander Sidorov, \and
Viswanath Sivakumar,
Andrew Tulloch,
Xiaodong Wang,
Yiming Wu,
Hector Yuen,
Utku Diril, \and
Dmytro Dzhulgakov,
Kim Hazelwood,
Bill Jia,
Yangqing Jia,
Lin Qiao,
Vijay Rao,
Nadav Rotem, \and
Sungjoo Yoo and
Mikhail Smelyanskiy\\
Facebook, 1 Hacker Way, Menlo Park, CA
}
\maketitle

\thispagestyle{empty}

\begin{abstract}

The application of deep learning techniques resulted in remarkable improvement of machine learning models. In this paper we provide detailed characterizations of deep learning models used in many Facebook social network services. We present computational characteristics of our models, describe high-performance optimizations targeting existing systems, point out their limitations and make suggestions for the future general-purpose/accelerated inference hardware. Also, we highlight the need for better co-design of algorithms, numerics and computing platforms to address the challenges of workloads often run in data centers. 

\end{abstract}

\section{Introduction}

Machine learning (ML), deep learning (DL) in particular, is used across many social network services. The high quality visual, speech, and language DL models {\em must scale to billions of users of Facebook's social network services}~\cite{hazelwood2018applied}. 

\begin{figure}
    \centering
    \includegraphics[width=0.90\columnwidth]{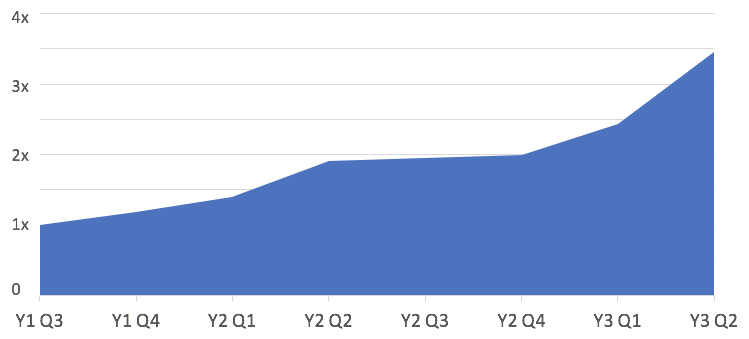}
    \caption{Server demand for DL inference across data centers}
    \label{fig:capacity}
\end{figure}

The power consumption in data centers\footnote{The collective power consumption of data centers around the world would be ranked 4th behind only China, US and EU~\cite{avgerinou2017trends}} used to run these models has been rapidly increasing over time. A significant fraction of the future demand is expected to come from workloads corresponding to DL inference, as shown on Figure~\ref{fig:capacity}. The higher DL inference demand is due to the expanding range of corresponding applications and the steady improvement in the quality of DL models, which is often associated with the increase in compute and memory requirements~\cite{ai_and_compute}. 

In order to tackle this trend, a lot of research has been done on optimizing computing platforms for DL, including but not limited to~\cite{jouppi2017datacenter,brainwave,parashar2017scnn,aklaghi2018snapea,sharma2017bit,outlier,scalable_tops,hazelwood2018applied}.
However, a great challenge has been the fast pace of changes in DL applications. For instance, the previously relevant AlexNet~\cite{krizhevsky2012imagenet} is no longer representative of the computation characteristics of today's computer vision (CV) systems. The rate of change in DL models is so fast that hardware optimized for old models can easily become inefficient for new models.

In order to perform a characterizations of the DL models and address aforementioned concerns, we had direct access to the current systems as well as applications projected to drive them in the future. Many inference workloads need flexibility, availability and low latency provided by CPUs. Therefore, our optimizations were mostly targeted for these general purpose processors. However, our characterization suggests the following general requirements for new DL hardware designs:

\begin{itemize}
    \item High memory bandwidth and capacity for embeddings
    \item Support for powerful matrix and vector engines 
    \item Large on-chip memory for inference with small batches 
    \item Support for half-precision floating-point computation
\end{itemize}

These requirements result from the characteristics of DL models important to us now (and projected to be in the future), our experience in optimizing DL applications for current computing platforms as well as their limitations found from our experiences. In particular, we highlight a gap in characteristics between the models commonly studied by the systems community and ones running in our data centers, with implications for future processor design.

\section{Characterization of DL Inference}
\label{sec:characterization}

This section highlights characteristics of DL inference workloads that are of interest in our data centers. Section~\ref{subsec:models} describes DL models used in our social network services and discusses trends observed in their evolution over time. Section~\ref{subsec:characteristics} presents detailed characteristics, focusing on aspects related to processor architecture design, and Section~\ref{subsec:kernel} details their common computational kernels.

\subsection{Representative Models}
\label{subsec:models}

We divide inference workloads into three categories.
The first provides personalized feed, ranking or recommendations, based on previous user interactions. The second and third are used for content understanding, visual and natural language content, respectively. The latter infer information used for powering recommendations, integrity and security such as detecting objectionable content.

\subsubsection{Ranking and Recommendation}
\ \\
Recommendation systems are one of the most common DL workloads in data centers with many applications like ads, feed, and search. Recommendation is usually formulated as an event-probability prediction problem, where an ML model predicts the probability of one or multiple events at the same time. The items associated with the most likely events are ranked higher and shown to the user ~\cite{he2014practical}.

Without going into a comprehensive scientific literature review, we point out that over time the ML models and recommendation systems have evolved to incorporate neural networks (NNs). The latter has progressed from matrix and tensor-based factorizations~\cite{frolov2017tensor,koren2009matrix} to autoencoder and neural collaborative filtering~\cite{he2017neural, kuchaiev2017training,sedhain2015autorec}. Further advances led to the development of more complex models, such as wide and deep as well as deep cross neural networks, which have been successfully applied in practice~\cite{covington2016deep,he2016deep,wang2017deep,zhou2018deep}. 

\begin{figure}
    \centering
    \includegraphics[width=0.8\columnwidth]{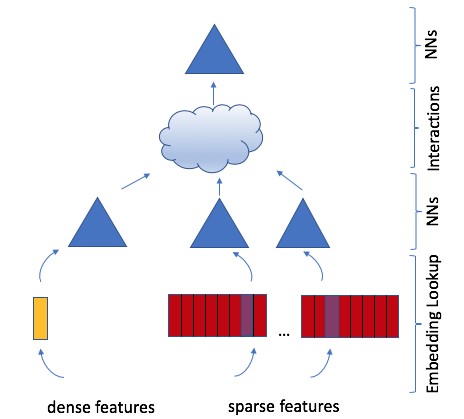}
    \caption{A deep learning recommendation model}
    \label{fig:ranking_model}
\end{figure}

These models usually use a combination of signals from dense and sparse features. The former are represented as a vector of real values, while the latter are often represented as indices of an one-hot encoded vector in a high-dimensional space. The sparse features are processed with embedding lookups that project sparse indices to a lower dimensional space. As in Figure~\ref{fig:ranking_model}, the resulting embeddings are combined with the dense features to produce higher order interactions, for example using a set of fully connected layers (FCs) or parameter-less additive and multiplicative mixing~\cite{rendle2010}.

The embedding tables can easily contain billions of parameters, while FCs usually have a modest number of parameters. The size of these models is often bound by the memory of the system at hand and can easily require a memory capacity exceeding tens of GBs.

These models often have to predict event-probabilities for multiple ad or feed candidates for a single user, usually within 100s ms time constraint. These properties allow us to leverage batching to achieve high performance in FCs. However, the overall model's execution tends to be memory bandwidth bound and is dominated by the embedding lookups. These look-ups perform a large number of mostly random accesses across table columns, but read an entire column vector for each such random access. For more details, refer to SparseLengthsSum operator in Caffe2.

\noindent
{\bf Future Trends:}
\begin{itemize}

\item
Model Exploration: recent studies explore explicitly incorporating time into the event-probability models~\cite{beutel2018latent,wu2017recurrent}. We believe that such techniques will lead to better models in the future but require more compute demand.

\item
Larger Embeddings: Adding more sparse signals and increasing embedding dimensions tends to improve model quality. Therefore, we expect even larger embeddings to be used. This will further increase the pressure on memory and leads to systems with larger memory capacity, while putting more focus on distributed training and inference.

\end{itemize}

\subsubsection{Computer Vision}
\ \\
CV models were some of the earliest to adopt DL techniques. They rely on convolutions that apply C$_\text{i}$ $\times$ K$_1 \times $ K$_2$ filters on the 
B $\times$ [F $\times$] C$_\text{i}$ $\times$ height (H) $\times$ width (W) input images with B batch size and C$_\text{i}$ input channels or video clip with F frames and produce a result with C$_\text{o}$ output channels.

\ \\
{\bf Image Classification} involves matching images to classes. Currently, ResNets are widely used for classification ~\cite{he2016deep}. However, recently much larger ResNeXt models have shown state-of-the-art accuracy even with weakly supervised training on billions of Instagram images~\cite{mahajan2018exploring,xie2017aggregated}.  For example,  our ResNeXt-101-32x4d model contains 43M parameters and requires 8B multiply-add operations during inference, relying on group convolutions with G=32 and d=4\footnote{In a group convolution, only the input channels in the same group are used for computing a given output channel. A group convolution with total C$_\text{i}$ input, C$_\text{o}$ output channels and G groups is essentially G independent convolutions each with d=C$_\text{i}$/G input and C$_\text{o}$/G output channels. A special case where C$_\text{i}$=C$_\text{o}$=G and consequently group size d=1 is called depth-wise convolution.
} in its first residual block. The largest configuration ResNeXt-101-32x48d contains 829M parameters and requires 153B multiply-add operations during inference, relying on group convolutions with G=32 and d=48 in its first residual block. It further improves the Top-1 validation accuracy of ImageNet-1K by 4\% to 85.4\%~\cite{mahajan2018exploring}.

\noindent
{\bf Object Detection} involves identifying specific regions that contain objects of interest. One of the largest scale object detection systems running in data centers today is the text detection part of the Rosetta system used to understand text in images~\cite{borisyuk2018rosetta}. It uses the Faster-RCNN-Shuffle model, that relies on Faster-RCNN architecture~\cite{ren2015faster}, with the ResNet trunk replaced with ShuffleNet~\cite{shufflenet}, which uses 3$\times$3 depth-wise convolutions and 1$\times$1 group convolutions with d=4. Despite ShuffleNet efficiency, object detection tends to be more time consuming than classification for the following reasons.

First, detection requires high resolution for accuracy. Whereas 224$\times$224 images are sufficient for classification, detection typically resizes images such that the maximum side is 800 while maintaining the aspect ratio. Therefore, a typical input size of dimension 3$\times$800$\times$600 for object detection is 9.5$\times$ larger than a typical input for classification.

Second, Faster-RCNN employs a region-proposal based approach where the final convolutional block is batched over many proposals of smaller spatial resolution. In Rosetta, the activations tend to be of dimensions [25-100 proposals] $\times$ [544 or 1088 channels] $\times$ [7,14] $\times$ [7,14]. The spatial resolution is typically 7$\times$7 or 14$\times$4, with large number of channels. Hence the number of proposals is a limiting factor in the number of objects that can be detected and is typically bounded due to computational cost.

\noindent
{\bf Video Understanding} historically has taken frame-based approach where sampled video frames are applied through image models. However, recently 3D convolutions gained wide adoption owing to higher accuracies given their ability to model temporal in addition to spatial domain~\cite{tran2015learning}. Extensive studies have been done to analyze the performance vs. accuracy trade-off of vanilla 3D ResNets compared to factorized 3D convolutions as in Res(2+1)D~\cite{tran2018closer}, ResNeXt-3D and ShuffleNet-3D~\cite{resnext3d}. In particular, ResNeXt-3D with depth-wise convolutions, which factorizes the 3D convolution into channel and spatiotemporal dimension, requires 3$\times$ less FLOPs than Res(2+1)D, which factorizes the 3D convolution across spatial and temporal dimension. Further, trading off spatial resolution for increasing clip length shows improved accuracy. In the future, increasing both the temporal and spatial resolution would be important for more complex video understanding tasks, such as object detection or action recognition.

\noindent
{\bf Future Trends:}
\begin{itemize}
\item{Model Exploration:} There is an increasing trend to fine-tune the last few layers of a model specific to each application (such as adding additional categories for classification) while all applications share a common trunk. This leads to the inference cost per image increasing linearly as a factor of the computational cost of only the final few layers. These final convolutions typically have a large number of channels and work on much smaller spatial resolutions, which can be important optimization targets.

\item{Convolution Types:} Group/depth-wise convolutions such as in ResNeXt and ShuffleNet, originally introduced for mobile inference, have increasingly been adopted in the data center due to accuracy and FLOP efficiency. Depth-wise convolutions are memory bandwidth bound, while majority of FLOPs spent elsewhere: e.g. ResNeXt-3D has 97.1\% of all FLOPs in 1$\times$1$\times$1 convolutions.

\item{Large Activations:} Image understanding is moving beyond simple classification tasks into more complex domains such as object detection, action recognition, and human pose estimation, which requires larger activations. For instance, tasks like object detection require higher resolution for accuracy, and video inference with more frames per clip demonstrates higher accuracy due to more temporal context. More CV tasks will see this trend, adding pressure on on-chip memory capacity and/or off-chip memory bandwidth.
\item{Batch Size:} Although CV inference typically does not have very strict latency requirements, small batches are still preferable in non-classification use cases.
Whereas classification tasks perform well when the aspect ratios are distorted into a fixed shape like 224$\times$224, doing so results in huge accuracy drops in complex tasks like object detection, making batching difficult.
Moreover, owing to large activations, increasing batch size puts further pressure on on-chip memory capacity.

\end{itemize}

\subsubsection{Language Models}

Neural machine translation (NMT) has become the dominant approach to machine translation~\cite{kalchbrenner2013recurrent,Neubig2017,sutskever2014sequence,bahdanau2014neural}. It relies on the encoder-decoder approach, also called seq2seq. The former encodes the input sentence, while the latter decodes the encoding into the target output sentence.

This approach has been successfully applied to many natural language processing tasks, such as summarization~\cite{nallapati2016abstractive,fan2017controllable}, speech recognition~\cite{graves2013speech}, syntactic and semantic parsing~\cite{cheng-EtAl:2017:Long,vinyals2015grammar} as well as question answering and dialog systems~\cite{nmt_qa,nmt_dialog}.

Encoder-decoder approaches vary according to the encoder and decoder implementation. A major challenge with NMT is the dependence of a translation segment on its position in the sentence. This motivated the reliance on recurrent neural networks (RNNs) as one can encode the statement's position in the recurrent network during translation.
This approach has shown successful results and is widely used in practice~\cite{sutskever2014sequence,bahdanau2014neural}.
In this approach, the encoder and the decoder are typically implemented using a Gated Recurrent Unit (GRU)~\cite{cho2014learning} or a Long Short Term Memory (LSTM) cells~\cite{hochreiter1997long}.

\begin{table*}
\footnotesize
    \centering
    \begin{tabular}{p{0.10\textwidth}|p{0.15\textwidth}|p{0.08\textwidth}|p{0.08\textwidth}|p{0.08\textwidth}|p{0.12\textwidth}|p{0.11\textwidth}|p{0.08\textwidth}}
\Xhline{2\arrayrulewidth}
Category & Model Types & Model Size (\# params) & Batch Size (typical) & Max. Live Activations & Arith. intensity (weights) & Arith. intensity (act. \& weights) & Latency (constraints) \\
\Xhline{2\arrayrulewidth}
\multirow{2}{*}{Recommendation} & FCs       & 1--10M        & 1--100 & >10K & 20--200 & 20--200 & 10s of ms \\
\cline{2-8}
 &                          Embeddings & >10 Billion & 1--100 & >10K   & 1--2    & 1--2    & 10s of ms   \\
\Xhline{2\arrayrulewidth}
\multirow{4}{*}{\shortstack{Computer\\ Vision}} & \multirow{1}{*}{ResNet-50} & 25M      & 1 image     & 2M   & avg. 303/min. 100 & avg. 164/min. 25 & \multirow{4}{*}{\shortstack{No strict\\ constraints}} \\
\cline{2-7}
& \multirow{1}{*}{ResNeXt-101-32x4-48} & 43--829M & 1 image & 2.4--29M & avg. 380/min. 100 & avg. 188/min. 28 & \\
\cline{2-7}
& \multirow{1}{*}{\shortstack{Faster-RCNN-Shuffle}} & 6M & 1 image & 13.2M & avg.3.5K/min.2.5K & avg. 145/min. 4 & \\
\cline{2-7}
& \multirow{1}{*}{ResNeXt3D-101} & 21M & \multirow{1}{*}{\shortstack{1 clip}} & 58M & avg. 22K/min. 2K & avg. 172/min. 6 & \\
\Xhline{2\arrayrulewidth}
Language & seq2seq (GRU/LSTM) & 100M-1B & 1-8 tokens & >100K & 2--20 & 2--20 & 10s of ms \\
\Xhline{2\arrayrulewidth}
    \end{tabular}
    \caption{Resource requirements of representative DL inference workloads implemented on CPU. The batch size can often be increased with more compute throughput, while meeting latency requirements. We point out that 1 clip consists of 8-16 frames.}
    \label{tab:characteristic}
\end{table*}

\noindent
{\bf Future Trends:}
\begin{itemize}
\item
Model Exploration:
Results have shown that adding more layers and ensembles improves translation quality, but leads to larger NMT models~\cite{sutskever2014sequence}. Reranking the output of a model is a successful approach that can be used together with ensembles~\cite{sennrich-haddow-birch:2016:WMT}. Also, multilingual models are an attractive way to scale one model to multiple languages but each multilingual model may need more capacity to handle multiple language pairs~\cite{johnson2016google,schwenk2017learning}. 

\item
Parallelism:
While successful, RNN-based approaches impose dependencies on each translated word, making it difficult to utilize highly parallel computing architectures such as GPUs.
Recognizing this has motivated NMT models that lift the time dependencies imposed by RNNs. In \cite{gehring2017convolutional}, both the encoder and decoder are implemented as stacked convolutions. In \cite{vaswani2017attention}, the transformer model is introduced which removes the need for recurrence or convolution altogether and instead only relies on the attention mechanism to improve achievable hardware parallelism at the expense of additional computation. Results from this work show that NMT training time can be significantly reduced while having the additional benefit of model generality.
While these approaches benefit from improved parallelism in both the encoder and the decoder during training and the encoder during inference, a typical inference generates an output sequentially using beam search. A more recent work has attempted to remove the time dependency in the decoder at inference time ~\cite{gu2017non}.

\item
Batch Size:
Inference with small batches is well suited in the context of instant translation. However, large-scale inference can also be done offline for localization purposes. In that case, using larger batch sizes can be beneficial as throughput becomes more important than latency.

\end{itemize}

\subsection{Compute Characteristics}
\label{subsec:characteristics}

Let the arithmetic intensity be defined as (\# of operations needed to evaluate) / (\# of elements incurred in data traffic) during the execution of model. The compute, memory capacity, and memory bandwidth demand of our representative DL inference workloads is shown in Table~\ref{tab:characteristic}. We report two arithmetic intensities: (i) assuming only weights are incurring the traffic, for example when all activations fit in a level closer to compute in the memory hierarchy, and (ii) assuming that both weights and activations are incurring traffic.

For DL hardware designs, there are notable differences between DL workloads found in our representative sample and those commonly studied in the systems community.

First, {\em embedding layers stand out with huge model sizes (more than 10s of GBs) and significantly low arithmetic intensities}.
Mathematically, the operation we perform on the embedding tables is a sparse-matrix times dense-matrix multiplication.
The sparse matrix has >10 rows and >10M columns, each row with >10 non-zeros.
This matrix is multiplied with a dense matrix with >10M rows and >10 columns.

The embedding lookup operation can be an interesting opportunity for applying emerging memory technologies and specialized hardware. On one hand, more expensive High-bandwidth memory (HBM) could be useful because it provides higher bandwidth but unfortunately its current capacity is limited. On the other hand, Non-volatile memory (NVM) is an economical alternative to store embeddings compared to DRAM, but the associated bandwidth is too low to be practical out of the box. Further, the memory access pattern to embedding tables has low temporal locality which makes caching challenging, while low spatial locality often results in underutilization (due to access granularity of 10s of Bytes versus NVM block size). Nonetheless, several techniques have been proposed to mitigate these problems~\cite{Assaf2018}.

\begin{figure}
    \centering
    \includegraphics[width=\columnwidth]{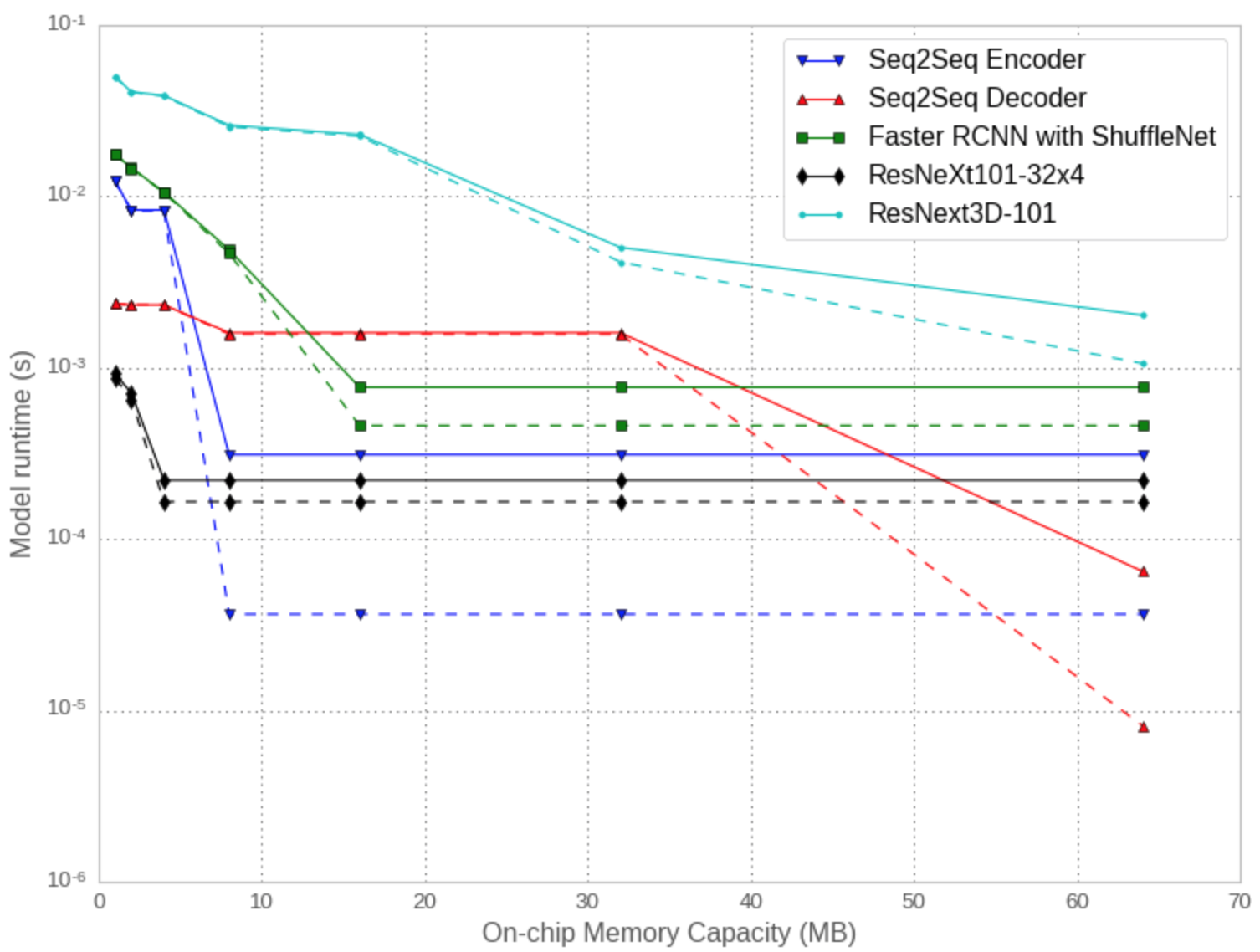}
    \caption{Runtime roofline analysis of different DL models with parameters stored as int8 numbers on a hypothetical accelerator with 100 TOP/s and 100 GB/s DRAM bandwidth. The performance is shown for varying on-chip memory capacity with 1 TB/s (solid) and 10 TB/s (dashed) bandwidth.}
    \label{fig:sram_analysis}
\end{figure}

Second, {\em recent models can benefit more from larger on-chip memory capacity}.
In a hypothetical accelerator with 100 TOP/s and 100 GB/s DRAM bandwidth the performance projected by a roofline model\footnote{We assume that the model parameters are stored as int8 numbers. We apply a roofline model for each layer, where each layer differs in whether it reads activations/weights from off- or on-chip memory based on a simple greedy on-chip memory allocation algorithm ~\cite{williams2009roofline}.} improves with larger on-chip memory capacities, as shown in Figure~\ref{fig:sram_analysis}.
This is not only driven by larger models, such as NMT seq2seq and ResNeXt-101, but also by larger activations, such as 800$\times$600 input images for ShuffleNet and videos for ResNeXt-3D.

Notice that the FC layers in recommendation and NMT models use small batch sizes so performance is bound by off-chip memory bandwidth unless parameters can fit on-chip. The batch size can be increased while maintaining latency with higher compute throughput of accelerators~\cite{jouppi2017datacenter}, but only up to a point due to other application requirements.
The number of operations per weight in CV models are generally high, but the number of operations per activation is not as high (some layers in the ShuffleNet and ResNeXt-3D models are as low as 4 or 6).
This is why the performance of ShuffleNet and ResNeXt-3D varies considerably with on-chip memory bandwidth as shown in Figure~\ref{fig:sram_analysis}.
Had we only considered their minimum 2K operations per weight, we would expect that 1 TB/s of on-chip memory is sufficient to saturate the peak 100 TOP/s compute throughput of the hypothetical accelerator.  As the application would be compute bound with 1 TB/s of on-chip memory bandwidth, we would expect there to be no performance difference between 1 TB/s and 10 TB/s.

Third, {\em common primitive operations are not just canonical multiplications of square matrices, but often involve tall-and-skinny matrices or vectors}. These shapes arise from group/depth-wise convolutions that have recently become popular in CV, and from small batch sizes in Recommendation/NMT models due to their latency constraints. Therefore, it is desired to have a combination of matrix-matrix engines to execute the bulk of FLOPs from compute-intensive models in an energy-efficient manner and powerful enough vector engines to handle the rest.

\subsection{Computation Kernels}
\label{subsec:kernel}

\begin{figure}
    \centering
    \includegraphics[width=\columnwidth]{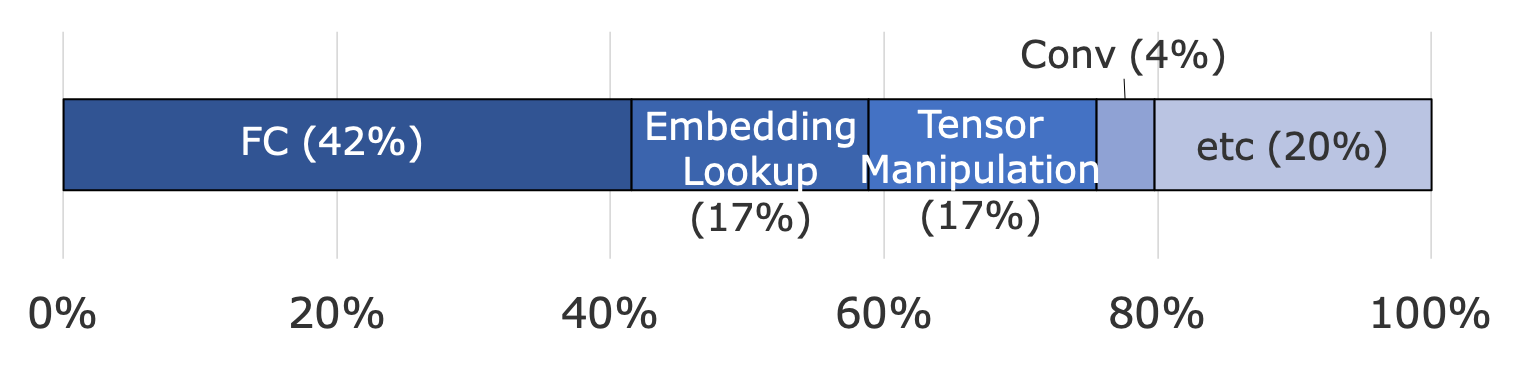}
    \caption{Time spent in Caffe2 operators in our data centers. }
    \label{fig:breakdown}
\end{figure}

Let us now illustrate the time spent in different computational kernels on CPU in our data centers.
Figure~\ref{fig:breakdown} shows that FCs are the most time
consuming operations, followed by embedding lookups and tensor manipulations\footnote{
``Tensor Manipulation'' refers to concatenation (for combining dense and sparse features in Figure~\ref{fig:ranking_model}), splitting, slicing, and so on, which are good targets for whole graph optimizations discussed in Section~\ref{subsec:whole_graph}.}.

\begin{figure}
    \centering
\subfloat[Activation]{
\includegraphics[width=0.9\columnwidth]{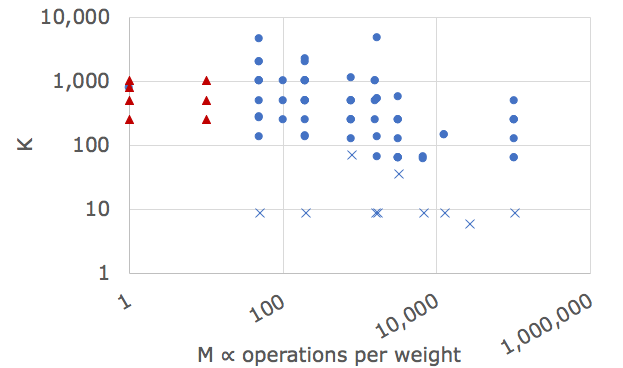}
} \\
\subfloat[Weight]{
\includegraphics[width=0.9\columnwidth]{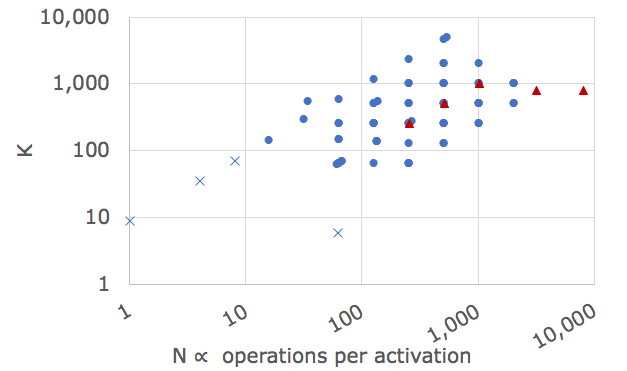}
}
\caption{Common activation and weight matrix shapes, with {\footnotesize $\bigtriangleup$}: FCs, $\times$: group and depth-wise convolutions, {\scriptsize $\bigcirc$}: other ops}
\label{fig:matrix_shapes}
\end{figure}

Following Caffe2 framework convention, the FC operator is defined as $\mathbf{XW}^T$, with $M \times K$ matrix $\mathbf{X}$ and $N \times K$ matrix $\mathbf{W}$ as inputs, and K being the inner reduction dimension. The convolutions can be logically transformed to matrix multiplications using {\tt im2col}, which results in $M=BHW$, $N=C_\text{o}$ and $K=C_\text{i}K_1 K_2$ as shown in Figure~\ref{fig:matrix_shapes}. 

We often refer to the number of rows $M$ as effective batch size or batch/spatial dimension, while $N$ as the output feature dimension. If $M$ or $N$ are small (e.g., {\scriptsize $\bigtriangleup$} and $\times$ corresponding to FC and group/depth-wise convolutions), the matrix-matrix multiplication becomes narrow and more closely resembles matrix-vector multiplication, with performance deteriorating accordingly from BLAS3 to BLAS2 levels. In this case a matrix-matrix multiplication engine is expected to have a low utilization. This happens for small batch sizes (e.g., recommendation and NMT) and group convolution with few output channels per group (e.g., ShuffleNet). 

The number of operations per weight read is proportional to the batch/spatial dimension, and the number of operations per activation read is proportional to the output feature dimension. If either is small, then performance is expected to be bound by memory bandwidth.
When an $M \times K$ matrix activation matrix is multiplied with a $K \times N$ weight matrix, we compute $2MKN$ operations while reading $KN$ weights, leading to $2M$ operations per weight.
For example, when a batch/spatial dimension ($M$) is 10, the operations per weight is 20.
In this case, if model parameters are stored as int8 numbers then saturating 100 TOP/s architecture would require 5 TB/s of memory bandwidth for reading weights.
Similarly, the number of operations per activation is $2N$.
With an output feature dimension of 8, operations per activation of 16 would require 6.25 TB/s for reading input activations.

Note that the overall arithmetic intensity of a DL model can be misleading and we should also look at its individual layers. For example, even though the depth-wise convolutions in ShuffleNet and ResNeXt account for only 2\% of total FLOPs, if a hypothetical accelerator can achieve 100 TOP/s for the other convolutions and only 2 TOP/s for the depth-wise convolutions due to memory bandwidth limitations, time spent in the depth-wise convolutions will be comparable to the others.

Finally, we point out that standard benchmarks, like DeepBench~\cite{deepbench}, typically give more emphasis on batch sizes larger than what is encountered in our use cases. They do not capture small reduction dimensions in depth-wise convolutions, and big activation tensors in image detection and video models.

\section{Performance Optimizations}
\label{sec:optimization}

DL inference workloads running on Facebook's social network services need to serve billions of users with fluctuating capacity demand over time~\cite{hazelwood2018applied}. Therefore, the availability and flexibility of computing resources is important. In addition, many inference workloads require low latency. These are the reasons why, currently, most inference workloads run on CPUs. Even though accelerators can significantly reduce inference time and improve energy efficiency, CPUs will still be responsible for a good portion of DL inference, especially in cases where tight integration with business logic is needed.

\subsection{Data Center Fleet-wide DL Inference Profiling}

Our data-centers are running diverse DL inference workloads. Table~\ref{tab:characteristic} lists representative models, but by no means covers all of our models and new models with new types of data and varying tensor shapes are always coming online. Therefore, {\em it is important to continously monitor DL model performance characteristics fleet wide}. DL operations typically utilize a large fraction of peak compute or memory bandwidth,  depending on their arithmetic intensity, and are less limited by memory latency or control overheads compared to other typical data center workloads. They often involve regular compute and memory access patterns, lending themselves as good targets of analytical performance models. 

For this purpose we have implemented the observer software design pattern that can be applied to individual operators and are executed at the start and end of the operator.  We have developed a number of functions called by observers that track performance metrics for each operator's execution (refer to Caffe2 operator cost inference functions for more details).  When considered in conjunction with the layer's full specification such as layer type, input/output tensor shapes, and element types, we can understand whether a given layer execution should be memory-bandwidth or compute bound.  Viewed differently, we can estimate the benefits of optimizing any specific operator.  This is particularly useful as it gives us the necessary data to estimate the priority of a considered optimization.  

In order to keep track of the accuracy and identify inefficiencies in the roofline models we maintain detailed per-layer logs that measure execution time, memory bandwidth in GB/s and actual attained FLOP/s that are derived from hardware performance counters for sampled DL operator executions. A telemetry agent running on each host collects and compares this information with given predictions across all of our data centers. Also, to set realistic goals for our optimization efforts, we developed a number of benchmarks tuned for each potential bottleneck.

\subsection{Reduced Precision Inference}

The reduced-precision inference has been shown to be effective at improving compute throughput within a power budget, especially in mobile platforms. However, applying reduced-precision inference in data centers is nontrivial. 

First, while mobile platforms have widely adopted CV models such as ShuffleNet and MobileNet that trade-off accuracy for significant reduction in compute requirements~\cite{shufflenet, howard2017mobilenets}, DL inference in data centers prefers accurate but compute intensive models like ResNet~\cite{he2016deep} and ResNeXt~\cite{xie2017aggregated}. In particular, when DL inference is related to core services like feed or integrity/security the accuracy loss should be very small. Usually $<$1\% change in the accuracy compared with single-precision floating-point results is acceptable.

Also, while general purpose CPUs have high availability in data-centers, they have not yet adapted to rapidly increasing compute demand of DL inference and hence lack good support for high-performance reduced-precision inference. This is exacerbated by less mature high-performance and high-accuracy reduced-precision linear algebra libraries for CPUs compared to their higher precision counter parts.

\subsubsection{Performance Challenges}

Current generations of x86 processors~\cite{ia64} provide conversion instructions between half- and single-precision floating point numbers ({\tt vcvtph2ps} and {\tt vcvtps2ph}), but without native half-float (fp16) computation.
They also require a sequence of instructions ({\tt vpmaddubsw} + {\tt vpmaddwd} + {\tt vpadd}) to implement 8-bit integer multiplications with 32-bit accumulation with marginally higher ($\sim$33\%) compute throughput than that of single-precision floating point (fp32)~\cite{mkldnn-i8}.
The compute throughput of 8-bit integer multiplications with 16-bit accumulation can be about twice higher than fp32, but this often results in significant accuracy drops unless combined with outlier-aware quantization that will be described shortly.
On the other hand, VNNI instructions provide higher throughput int8 multiplications with 32-bit accumulation but they are not available in current x86 microarchitectures~\cite{cascade_lake}. As a result, we had to tailor optimization strategies based on the performance bottleneck. 

\begin{figure}
\centering
\subfloat[FC]{
\includegraphics[width=0.9\columnwidth]{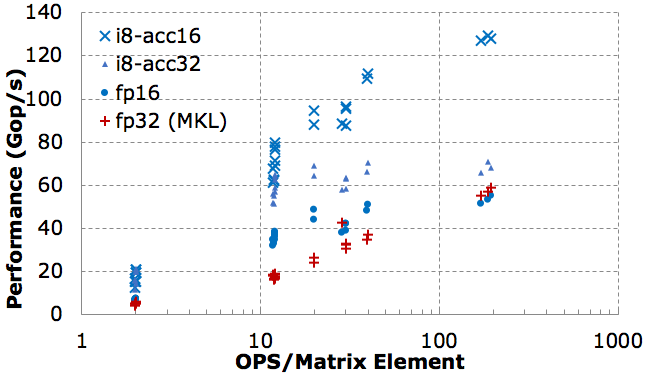}
}
\\
\subfloat[Conv]{
\centering
\includegraphics[width=0.9\columnwidth]{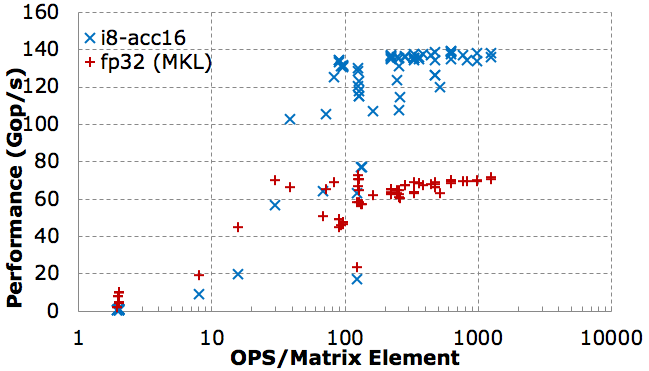}
}
\caption{Performance of FBGEMM in Gop/s vs. arithmetic intensity (2NMK/(NK + MK)) for multiplications of $M\times K$ and $K\times N$ matrices, compared with MKL GEMM in fp32.
}
\label{fig:fbgemm}
\end{figure}

If the performance is memory-bandwidth bound, then using fp16 when storing weights or using 8-bit multiplications with 32-bit accumulation (i8-acc32) can increase the arithmetic intensity by up to a factor of 2$\times$ and 4$\times$, respectively. In this case, we can obtain speedups proportional to the memory bandwidth saving, even when we save nothing with respect to the number of instructions. For example this happens in FCs with small batch sizes and group convolutions with a small number of channels per group (the extreme case being depth-wise convolution with just one channel per group).

We have designed and implemented a reduced-precision linear algebra library for DL inference called FBGEMM~\cite{fbgemm, fbgemm_blog}.
Figure~\ref{fig:fbgemm}(a) plots the performance of our optimized fp16 and i8-acc32 matrix multiplication (GEMM) in FBGEMM compared with Intel MKL's fp32 GEMM. The experiments are performed on a single thread running on Intel Xeon E5-2680 v4 with turbo mode off using Intel MKL version 2017 update 3. Notice that for cases with low arithmetic intensity our fp16 and i8-acc32 GEMMs obtain up to 2$\times$ and 4$\times$ speedups over MKL's fp32 GEMM, respectively. For instance, applying our fp16 GEMM, we obtain up to 2$\times$ speedup in FC layers in a recommendation model with 15\% overall latency reduction. Also, applying our i8-acc32 GEMM, we obtain overall 2.4$\times$ speedup in the Faster-RCNN-Shuffle used for our optical character recognition application.

If the performance is bound by the instruction throughput, then we use 8-bit multiplications with 16-bit accumulation and periodic spills to 32-bit accumulators (i8-acc16), which can provide $\sim$2$\times$ compute throughput over fp32. To avoid saturation and accuracy drops, we employ outlier-aware quantization that separates out weights with bigger magnitude as outliers ~\cite{outlier}.
Here, we consider a typical threshold for outliers, where a weight is not an outlier if representable with 7 bits (i.e. the value of weight is between -64 and 63).
We split the weight matrix into two parts, $\mathbf{W} = \mathbf{W}_{main} + \mathbf{W}_{outlier}$, where $\mathbf{W}_{main}$ is in 7 bits and $\mathbf{W}_{outlier}$ contains the residual.
The matrix multiplication, $\mathbf{X W}^\intercal$, is calculated in two stages, where $\mathbf{X W}_{main}^\intercal$ uses 16-bit accumulation, and $\mathbf{XW}_{outlier}^\intercal$ uses 32-bit accumulation.
We find that $\mathbf{W}_{outlier}$ becomes a sparse matrix, often with density less than 0.1\%, especially when combined with symmetric quantization~\cite{krishnamoorthi2018quantizing}.
Since $\mathbf{W}_{outlier}$ is sparse, $\mathbf{XW}_{outlier}^\intercal$ accounts for a small fraction of the total time.
Figure~\ref{fig:fbgemm}(b) plots the performance of our i8-acc16 GEMM compared with MKL GEMM in fp32, which achieves up to 2$\times$ speedup for matrix shapes with high arithmetic intensity.
In particular, applying our i8-acc16 GEMM to ResNet-50, we obtain 1.7$\times$ speedup over MKL in fp32.

Even though some of the applied optimizations are done to work around limitations of current x86 processors, they provide insight for future DL hardware optimizations.
Our optimizations show it is useful to apply different quantization techniques depending on where the performance bottleneck lies.
For example, quantization techniques that are primarily for saving storage and bandwidth should be tested with embedding layers, FCs with small batch size, and depth-wise convolutions.
Our experience with outlier-aware quantization shows that a high-performance sparse linear algebra engine will be helpful not only for pruned models but also for reducing required precision of non-pruned models.
For example, 6-bit quantized models can be computed in 4-bit for main values while the outlier part is computed with the 6-bit sparse engine.

\subsubsection{Accuracy Challenges:}

Impressive progress has been made in low-precision DL inference, some of which consider even ternary or binary quantization~\cite{rastegari2016xnor,zhou2016dorefa}. However, even 8-bit quantization has presented its own set of challenges to meet the accuracy requirements of our DL workloads in data centers. The following five techniques were effective at meeting the accuracy requirements:
\begin{enumerate}
    \item {\em Fine-grain Quantization}. Instead of having a single quantization parameter per tensor, applying quantization in a finer granularity is often required. Examples are per output feature quantization in FCs, per output channel quantization in convolutions, per group quantization in group convolutions, or per-entry quantization in embedding tables.
    \item {\em Quantization-aware Training}. We found that quantization-aware training for example using fake quantization is important for meeting the accuracy requirements. This aligns with a recent white paper~\cite{krishnamoorthi2018quantizing} that shows the importance of per-channel quantization and quantization-aware training in quantizing CNNs for mobile platforms.
    \item {\em Selective Quantization}. Unlike mobile platforms which can highly prefer end-to-end quantization, DL inference in data centers should be able to fall back to floating-point in accuracy-sensitive parts of DL models. We systematically profile errors introduced by quantization per layer and skip quantization when the error is too high. Examples include the first and last few layers of CNNs.
    \item {\em Outlier-aware Quantization}. In addition to the outlier-aware quantization technique described previously for 16-bit accumulation, we can take advantage of the fact that the range of values can be confined much more once outliers are ignored. For example, instead of quantizing a tensor $\mathbf{W}$ for the range [min($\mathbf{W}$), max($\mathbf{W}$)], we can quantize for a smaller range, such that the \href{https://github.com/pytorch/pytorch/blob/master/caffe2/quantization/server/norm_minimization.cc}{L2 norm of quantization error} is minimized with respect to the distribution of $A$ values.
    Unlike weight tensors, activation tensors are not constant, so
    we collect distribution of activation tensors by running with calibration inputs from the training data.
    \item {\em Net-aware Quantization}. We can often further reduce the range we're quantizing for based on neighboring operators. For example, if an operator is only followed by ReLU, we can narrow down the range by excluding negative values.
\end{enumerate}

For instance, using these techniques, \href{https://github.com/caffe2/models/tree/master/resnet50_quantized}{a ResNet-50 model with int8 quantization} (except softmax) achieves 75.6\% Top-1 and 92.8\% Top-5 accuracy for ImageNet-1K validation set~\cite{deng2009imagenet}, which corresponds to only 0.3\% Top-1 and 0.1\% Top-5 accuracy drop compared to the baseline fp32 model~\cite{goyal2017accurate}.

\subsubsection{Software Challenges:}

Linear algebra operations for machine learning inference require optimizations that are quite different from those for high-performance scientific computing (i.e. HPC). The standard BLAS interface cannot provide the desired performance for the matrix shapes that are common in DL inference. Since the compute requirement in DL is rapidly changing, it can be premature to attempt to standardize a new linear algebra interface for DL, but it worthwhile to discuss the associated requirements and challenges.

As shown in Figure~\ref{fig:matrix_shapes}, typical matrix shapes in DL inference are smaller and often tall and skinny, compared to those in typical HPC applications. High-performance matrix multiplications often “pack” a block of input matrices into a format friendly for vectorization and cache locality.
For large enough square matrices, the overhead of packing can be amortized inside a single matrix multiplication adhering to the standard BLAS interface.
However, for tall-skinny matrices, we need to amortize the packing overhead across multiple matrix multiplications for constant weight matrices which requires a new interface that accepts a custom pre-packed matrix.

A significant fraction of DL computation is not strictly matrix multiplication.
For example, the convolution operator in CNNs can be expressed as {\tt im2col} followed by matrix multiplication, but this often does not lead to the highest performance due to the duplication of input activations and the overhead of {\tt im2col}.
Therefore, it is important to promote convolution as a first-class citizen of the interface to enable the computation of direct convolutions without {\tt im2col}.
This will also enable algorithmic optimizations such as Winograd or FFT-based convolution as in cuDNN with automatic choice of the best algorithm for given tensor shapes.
The native convolution interface is particularly important for group convolution with only a few channels per group.
If we individually apply {\tt im2col} followed by GEMM for each group, the reduction dimension and the output feature dimension are too small for efficient vectorization and parallelization.
Note that even the FC layer cannot be implemented strictly with only a GEMM operation as it involves a bias term which should be fused with the main GEMM to save memory bandwidth.
It is also desirable to fuse other common operations such as ReLU.

Reduced-precision fixed-point computation requires additional steps such as handling non-zero offsets used in asymmetric quantization and rescaling 32-bit intermediate results of matrix multiplication, which should be fused with the main GEMM to save bandwidth.
Google's gemmlowp library~\cite{gemmlowp} provides a well-designed interface of fusing “output pipeline” with the main GEMM.
However, gemmlowp doesn't provide native convolution interface and is mostly optimized for ARM Neon and Intel x86 SSE4.1, not for AVX2 and AVX-512.

Intel MKL-DNN is another library that provides high performance implementations of DL primitives on CPU.
MKL-DNN implements advanced features such as Winograd convolution in int8. On the other hand, FBGEMM has features such as outlier-aware quantization.
Therefore, some of our DL inference applications use FBGEMM and some others use MKL-DNN, depending on compute characteristics and operator availability.
Low-precision linear algebra for DL inference is still a young field, and we believe it is healthy to have multiple complementary solutions that can experiment different optimizations while adopting proven techniques from each other.

The below code snippet shows an example of our FBGEMM library interface. In this example, a templatized C++ function that performs a matrix multiplication for different data types is shown. The salient features of this interface are the way it accepts a packed B matrix (usually weights that can be packed once and used multiple times) and also a parameter for packing matrix A.
The packing of matrix A can be specialized and fused with memory bandwidth bound operations such as {\tt im2col}, row-wise sum for asymmetric quantization, or depth-wise convolution.
{\tt outProcess} parameter is templatized to support various types of data processing on output once the main GEMM part is finished  (similar to gemmlowp's output pipeline).
As previously mentioned, many matrices in DL inference are tall-skinny so the main kernels of matrix multiplication are dynamically generated just-in-time to take advantage of matrix size specific optimizations.
The FBGEMM library is \href{https://github.com/pytorch/fbgemm}{open source} and integrated with Caffe2 deeplearning framework. For more complete examples, refer to the tests and benchmarks in our open source project.

\begin{verbatim}
template<typename T_PACK_A, typename T_PACK_B,
         typename T_C, typename OUT_FUNCTOR>
void gemmPacked(
    // packed inputs
    T_PACK_A& packA, T_PACK_B& packedB,
    // output  
    T_C* C, uint32_t ldc,
    // post-processing functor, e.g. Relu
    OUT_FUNCTOR& outProcess);
\end{verbatim}

\subsection{Whole Graph Optimization}
\label{subsec:whole_graph}

While it is important to optimize the performance of individual operators as outlined in the previous subsections, we can get additional significant performance improvements by looking at the DL graph as a whole and performing cross-operation optimizations. A few different optimizations fall into this category, including operator fusion, data movement elimination, operator scheduling, and threading for both inter- and intra-op parallelism. This section focuses on operator fusion, specifically quantifying potential speedups of operator fusion. The realized speedup from operator fusion will heavily depend on the efficiency of underlying fused kernel.  Automatic generation of fused kernels is an active area of research and early productization efforts are underway~\cite{rotem2018glow, vasilache2018tensor, chen2018tvm, leary2017xla}. However, it is still often necessary to write fused kernels manually. For this reason, we focus our efforts in two directions: 1) to find the top few opportunities where we will get the most gains from fusion for our models that can be worth manual attention and 2) to find a broader set of opportunities for compiler generated kernels. 

Our approach to identify fusion opportunities for both cases is similar. We aim at identifying subgraphs that occur commonly in our workloads across the entire fleet; and are expected to have high speedup potentials. We log the complete graphs annotated with operator dependencies, frequency, and input/output tensor shapes. We then run a frequent subgraph mining algorithm on the nets captured. The idea here is to find all subgraphs that are executed frequent enough and order them on the basis of speedup potential from fusion. To perform the ordering, we use the input/output dimensions for the operators to compute a simple roofline model for the subgraph being considered. Specifically, we compute performance projected by the roofline model before and after fusion, and use the difference to estimate speedup potential. Note that we filter out some subgraphs based on specific operator pattern rules. For example, we rule out subgraphs with operators that are not data parallel and hence challenging to fuse. Finally, we run a top-k algorithm on the ordered subgraphs to return the top opportunities. 

With this analysis, we were able to find several opportunities for merging batched matrix multiplies with tensor manipulation operations. As analyzed in Figure~\ref{fig:breakdown}, these tensor manipulation operations comprise about 17\% of the overall DL inference CPU time. Most of these operations are memory bandwidth limited; merging them with compute bound operations resulted in a total of over 10\% savings in run time.

\section{Application Driven HW Co-design Directions}
\label{sec:implication}

This section discusses implications of the DL model characteristics and their optimization for software and hardware co-design. We believe that the server-side DL workload optimizations should be considered as a co-design problem along three axes: DL models, numerics (quantization, Winograd/FFT convolution, and sparsity), and hardware platforms. Also, the process should be driven by DL models because of their rapid changes and diversity. We highlight a few relevant observations in this regard next.

\noindent
{\bf Workload Diversity:} DL is a fast moving field while the design space of inference hardware is huge. Therefore, one needs a fast turn-around loop with performance modeling capability to predict benefits of various hardware and software co-optimizations based on workload characteristics captured from a wide range of up-to-date DL models. This study reveals the following characteristics of DL models running in our data centers. First, they have diverse compute patterns where matrices do not necessarily have “nice” square shapes. There are also many “long-tail” operators other than FC and convolutional layers. Therefore, in addition to matrix multiplication engines, hardware designers should consider general and powerful vector engines. Second, DL models in our data centers have diverse and sometimes conflicting demands to memory subsystem. For example, due to larger activation matrices or matrices with tall-and-skinny shapes, recent CV and NMT models need bigger on-chip memory capacity to sustain high compute throughput without being bottlenecked by off-chip memory bandwidth. However, we should not solely rely on on-chip capacity to fit the entire model because it is difficult to project on-chip memory capacity demand of future models. Some of our current recommendation models are already too big to fit on-chip memory. Recommendation models not only require a huge memory capacity but also high bandwidth. 

\noindent
{\bf Data Center Requirements:} When co-designing inference hardware for data centers, it is important to keep in mind that data center server-side DL inference has different requirements from mobile/embedded/IoT devices. For example, some quantization and pruning techniques report 2--3\% accuracy drops but that is often too high for data center environment and they are often not deployed. If quantization drops the accuracy of say 32x32d model by more than 1\% with less than 2$\times$ speedup, it can be more advantageous to just use the 32x16d model without quantization. In order to minimize accuracy drops from quantization, inference hardware for data centers should support per-channel quantization. They also should support fp16 or bfloat16 compute as a fallback in accuracy sensitive parts such as the last layer of some DL models.

\noindent
{\bf Service Dis-aggregation:} DL applications have distinctive compute and memory characteristics, compared to other typical data center workloads.
Specifically, DL inference often utilizes a higher fraction of peak FLOPs, memory capacity, and bandwidth.
As a result, other jobs on the same machine can suffer memory capacity and bandwidth pressure, or power limitation, e.g. reduction in turbo frequency when AVX2 or AVX-512 is used by deep learning workload~\cite{avx512}.
This reduces the performance of other important components such as business logic and has detrimental effect on full system performance.
Hence, a natural decision is to dis-aggregate DL inference into a separate tier (accelerated or not).
Dis-aggregation can also allow to pool requests from many front-end servers, increasing the batch size and hence compute efficiency.
A challenge is that inference queries and results need to be transferred between the tiers over the network.
Thus, the tier design, network bandwidth and latency, and compression techniques need to be carefully considered.
For example, a hypothetical accelerator with 100 TOP/s compute throughput would require a few GB/s PCIe and/or network bandwidth for the DL models listed in Table~\ref{tab:characteristic}, unless image decompression can be done within the accelerator or on the same host.

\noindent
{\bf DL Model and Hardware Co-design:}
It is important to co-design DL models to be aware of the cost associated with the required hardware resources.
While power/energy budget, on-chip memory capacity, and off-chip memory bandwidth are typically more scarce resources, research on efficient DL model architectures often only optimizes the number of floating-point operations.
When the performance is bandwidth bound, adding more FLOPs without increasing the bandwidth consumption can be a good way to improve the accuracy while maintaining the performance.
If adding 2$\times$ FLOPs to the FC part of a recommendation model and increasing the embedding dimension of its embedding table by 2$\times$ provide similar accuracy improvements, we would expect adding FLOPs will be the more economical direction.
Recovering accuracy losses from quantization by making DL models wider is an example of hardware cost aware trade-offs: int8 multiplication consumes more than 5$\times$ less energy and area compared to fp16 multiplication, hence there is a big room to recover the accuracy while maintaining the energy savings ~\cite{dally2015high,mishra2017wrpn}.
NMT models with higher arithmetic intensity and parallelism such as the transformer architecture also illustrate hardware cost aware trade-offs.

\section{Related Work}
\label{sec:related}

Recently, Hazelwood et al. presented a holistic characterization of ML workloads in data centers, covering inference, training, data acquisition and including a discussion of their diversity, huge data and compute capacity demands~\cite{hazelwood2018applied}. In contrast, our paper aims to provide insights that are useful for software/hardware co-design of DL applications, focusing on DL inference characteristics. 

Hardware accelerators for server ML workloads have been actively studied by academia and industry, with NVIDIA GPUs, Google TPUs and Microsoft Brainwave computing platforms being successfully used in data centers~\cite{Turing2018, jouppi2017datacenter, brainwave}. In particular, Google TPU relies on a systolic array accelerator mainly targeted for 8-bit matrix-matrix multiplication which is challenging to utilize for small batches and group/depth-wise convolutions. On the other hand, Microsoft Brainwave is a matrix-vector accelerator for low latency AI applications in data centers. It consists of dot-product engines which perform, with the broadcast vector and its local matrix weights, dot-product operations in parallel. The salient features of Brainwave are model pinning and block floating point representation. The large on-chip memory of FPGA is exploited to store weights on chip, avoiding off-chip DRAM accesses. Block floating point offers low precision computation by enabling 4- or 5-bit multiplications of mantissa and 5-bit additions of shared exponents. However, it is not clear if architectures like Brainwave are general enough to efficiently target our diverse DL inference workloads

Moreover, a number of techniques has been proposed to improve energy efficiency of DL inference by taking advantage of reduced precision and sparsity. NVIDIA’s SCNN skips computation with zero input in matrix multiplications, thereby offering significant improvements in energy efficiency~\cite{parashar2017scnn}. Akhlaghi et al. propose early stopping of convolution when the output is expected to be non-positive followed by ReLU~\cite{aklaghi2018snapea}. Sharma et al. present a systolic array accelerator called BitFusion which supports variable bit precision~\cite{sharma2017bit}. Park et al. present a zero-aware 4-bit accelerator called OLAccel which applies reduced precision to the majority of data while keeping a small fraction of large value data in high precision~\cite{outlier}, a technique also used in the optimizations described in this paper. Fleischer et al. propose a multi-TOP/s AI core supporting a wide range of precision from fp16 (for training) to 1- or 2-bit (for inference)~\cite{scalable_tops}.
Our paper shows that, while low-precision and sparse computation can significantly improve energy efficiency, they should meet the accuracy requirements of server-side DL inference to be widely used in data centers.

Finally, we point out that a number of DL benchmarks are actively being  developed~\cite{mlperf,nn_bench}. A benchmark framework has been presented where a model zoo of benchmark neural networks is provided and the performance of neural networks, optimized by users, is measured on real mobile devices remotely~\cite{nn_bench}. MLPerf aims at providing benchmarks for both server and mobile devices~\cite{mlperf}. These benchmarks will facilitate system-level performance measurements and comparisons on diverse software platforms like TensorFlow~\cite{leary2017xla} and PyTorch~\cite{paszke2017pytorch} as well as hardware architectures.

\section{Conclusion}
\label{sec:conclusion}

In the face of rapid innovation in deep learning and the increase of their computation and memory requirements, co-designing DL inference hardware for current and future DL models is an important but challenging problem.
We believe our DL inference characterization and optimization experience can provide useful insights for DL inference hardware designs.
We hope our paper can also contribute to discussion around software ecosystem such as benchmarking suites, linear algebra interface optimized for DL, and the compiler for optimizing and scheduling the whole graph of DL models, which are important parts of co-design process.

\section{Acknowledgements}
We would like to thank AML, Caffe2 and Glow team members for help with collecting information and reviewing this study. 

\bibliographystyle{plain}
\setlength{\bibsep}{0.0pt}
\footnotesize
\bibliography{references}
\end{document}